\documentclass{egpubl}
\usepackage{EGUKCGVC2023}
 
\WsPaper           

\usepackage[T1]{fontenc}
\usepackage{dfadobe}  

\usepackage{cite}  
\BibtexOrBiblatex
\electronicVersion
\PrintedOrElectronic
\ifpdf \usepackage[pdftex]{graphicx} \pdfcompresslevel=9
\else \usepackage[dvips]{graphicx} \fi

\usepackage{egweblnk} 

\usepackage{amsmath}
\usepackage{booktabs}

\title[Inpainting Normal Maps for Lightstage data]%
      {Inpainting Normal Maps for Lightstage data}

\author[H. Zuo \& B. Tiddeman]
{\parbox{\textwidth}{\centering Hancheng Zuo\thanks{First Author}$^{1}$\orcid{0009-0004-8029-974X}
        and Bernard Tiddeman\thanks{Corresponding Author}$^{1}$\orcid{0000-0001-7570-1192}
        }
        \\
{\parbox{\textwidth}{\centering $^1$Aberystwyth University, UK
       }
}
}

\begin{document}


\maketitle
\begin{abstract}
This paper presents a new method for inpainting of normal maps using a generative adversarial network (GAN) model. Normal maps can be acquired from a lightstage, and when used for performance capture, there is a risk of areas of the face being obscured by the movement (e.g. by arms, hair or props).  Inpainting aims to fill missing areas of an image with plausible data.  This work builds on previous work for general image inpainting, using a bow tie-like generator network and a discriminator network, and alternating training of the generator and discriminator.  The generator tries to sythesise images that match the ground truth, and that can also fool the discriminator that is classifying real vs processed images.  The discriminator is occasionally retrained to improve its performance at identifying the processed images.  In addition, our method takes into account the nature of the normal map data, and so requires modification to the loss function.  We replace a mean squared error loss with a cosine loss when training the generator.  Due to the small amount of available training data available, even when using synthetic datasets, we require significant augmentation, which also needs to take account of the particular nature of the input data.  Image flipping and in-plane rotations need to properly flip and rotate the normal vectors.  During training, we monitored key performance metrics including average loss, Structural Similarity Index Measure (SSIM), and Peak Signal-to-Noise Ratio (PSNR) of the generator, alongside average loss and accuracy of the discriminator. Our analysis reveals that the proposed model generates high-quality, realistic inpainted normal maps, demonstrating the potential for application to performance capture. The results of this investigation provide a baseline on which future researchers could build with more advanced networks and comparison with inpainting of the source images used to generate the normal maps.



\end{abstract}  
\section{Introduction}

Inpainting is a widely used image processing technique that involves filling in missing or damaged regions of an image. It has a wide range of applications, such as restoring damaged photographs \cite{bertalmio2000image}, removing unwanted objects from images \cite{criminisi2004region}, and improving image compression techniques \cite{mainberger2013edge}. In the field of medical imaging, inpainting has been used to reconstruct images from partial scans \cite{xiao2021mri}, while in the film industry, it has been applied for special effects and post-production work \cite{wexler2007space}.

Traditional inpainting techniques involve manually painting over missing areas with color or texture information from surrounding regions \cite{bertalmio2000image}. However, this process can be time-consuming and requires significant expertise \cite{telea2004image}.

In recent years, deep learning-based inpainting methods have emerged as a promising alternative, which have shown remarkable performance in generating high-quality inpainted images \cite{liu2018imageinpainting}. These methods use convolutional neural networks (CNNs) to learn the mapping between the input image and its missing regions, allowing them to fill in missing regions with visually appealing results.

This study focuses on developing an image inpainting system for normal map facial data acquired from a lightstage \cite{ma2007rapid, ghosh2011multiview}.  A lightstage uses a spherical arrangement of LED lights to provide a variety of illumination patterns.  Polarising filters on some of the LEDs and one of the cameras allow specular reflections to be eliminated from some images.  By capturing the specular and diffuse reflections under suitable lighting patterns, normal maps for surface and subsurface scattering can be estimated, along with the diffuse 'albedo' (i.e. shading free surface colours).  This allows realistic rendering of facial models under differently lighting.  Lightstage data has found various applications, including performance capture \cite{wilson2010temporal}, face analysis-by-synthesis \cite{smith2020morphable}, and facial image relighting \cite{legendre2020learning}.

The inpainting algorithm is inspired by the Context Encoders model proposed by Pathak et al. \cite{pathak2016context}, the bowtie-like model structure (encoder-decoder architecture) similar to U-Net proposed by Ronneberger et al. \cite{ronneberger2015unet}, and the Deep Convolutional Generative Adversarial Network (DCGAN) architecture proposed by Radford et al. \cite{Radford2015}. The Context Encoders model is a combination of deep CNNs and generative adversarial networks (GANs) \cite{Goodfellow2014}, which learns to inpaint missing regions of images in an unsupervised manner. The bowtie-like model structure inspires the design of the network layer structure for the generator and discriminator, whereas the DCGAN architecture serves as the foundation for designing the entire GAN model. The inpainting model uses contextual information present in the surrounding regions to generate a plausible reconstruction of the missing regions and is trained using a joint loss function that combines a reconstruction loss and an adversarial loss.

In this study, the goal is to adapt the existing inpainting system to work with lightstage data, in particular normal maps, directly.  This poses a number of challenges, including the lack of large sized datasets needed for training deep learning (hence requiring extensive augmentation), accounting for the particular form of the data when applying data augmentations such as flipping and rotation, and adapting the loss functions to properly account for the nature of the data.

This paper presents our model design, experimental results, and the conclusions drawn from these results, offering valuable insights for subsequent work in the field of deep learning-based image inpainting of normal maps.

\begin{figure*}[tbp]
    \centering
    \includegraphics[width=1.0\linewidth]{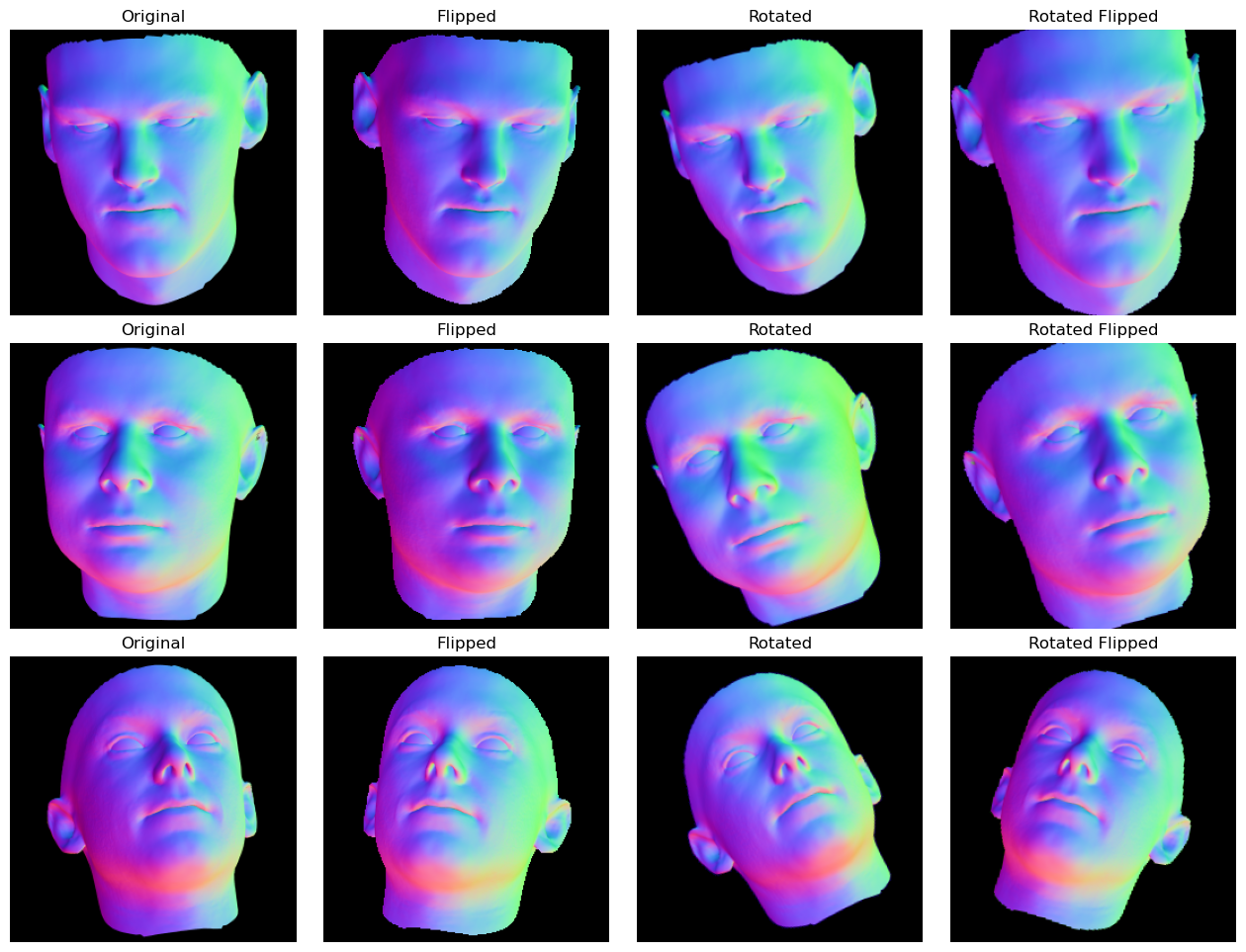}
    \caption{Example of augmentations, rows from left to right: Original; flipped; random zoom and rotation; flipped with (different) random zoom and rotation.}
    \label{fig:augmentation}
\end{figure*}

\section{Data Preparation}

Our work uses the SFSNet dataset \cite{sengupta2018sfsnet} of only 300 training and 105 testing normal map images.  This is a synthetic dataset generated from rendered 3D facial models.  Such a small training set is insufficient for effectively training a deep CNN, so data augmentation is required.  Flipping, rotating and zooming images are commonly used approaches to data augmentation, but care needs to be taken when dealing with normal map data.  

When flipping or rotating a standard RGB image the colours sampled from the input image location are written to the output location unchanged.  With normal maps, the sampled normal vector additionally needs to flipped or rotated to match the global image transformation.  To achieve this, we apply an additional matrix transform to the sampled normals.  

For the particular dataset used we also apply some additional operations so that the resulting images more closely resemble the unaugmented images.  The background values in the training set are allocated the RGB values (0,0,0), which becomes $(-1,-1,-1)/\sqrt(3)$ when converted to a normal vector.  When the normal map is flipped or rotated this vector is converted to a different direction.  We move these background vectors back to $(-1,-1,-1)/\sqrt(3)$ by identifying them using the (flipped / rotated / zoomed) mask to identify the background locations.  Points around the edge of the face model would often show artefacts where the uncorrected background was still showing, so we additionally erode the mask slightly to remove such artefacts.  Examples of original and augmented images are shown in figure~\ref{fig:augmentation}.  In this work we augment the dataset by (a) flipping all images, and (b) randomly rotating by $\pm20^o$ and zooming randomly by $\pm10\%$ resulting in $1200$ training normal map images in total.

In order to train the system random areas were masked using randomly generated irregular mask channels. The functions for generating mask channels were devised to accept parameters for the number of masks and mask dimensions. We train 3 systems using different masking styles: (a) random lines of varying thickness and directions, (b) randomly scattered small circles, and (c) a randomly located single large circluar area.  For the line masks, a 'size' variable is randomly generated that dictates the maximum line thickness in the mask, and used the cv2.line method from the OpenCV library to draw random lines, with intensity values representing occluded regions. A similar approach was used for generating the scattered circles and single circle masks.  The generated mask channels were normalized to maintain consistency with the pre-processed training data, aiding in simulating real-world occlusions.

A function performed element-wise multiplication of raw images and mask channels to create masked images with pixel values close to 0 in the masked areas, and original pixel values in the unmasked areas. The generated masks were applied to the original images using this function to remove the masked areas, and then the masked images were concatenated with the corresponding mask channels. This provided the necessary input data for training the generator to discern  missing image areas that required filling. This masked image data served as a learning base for the inpainting GAN model, enabling its ability to handle complex inpainting scenarios.

\section{Model Design}

\begin{figure*}[tbp]
    \centering
    \includegraphics[width=1.0\linewidth]{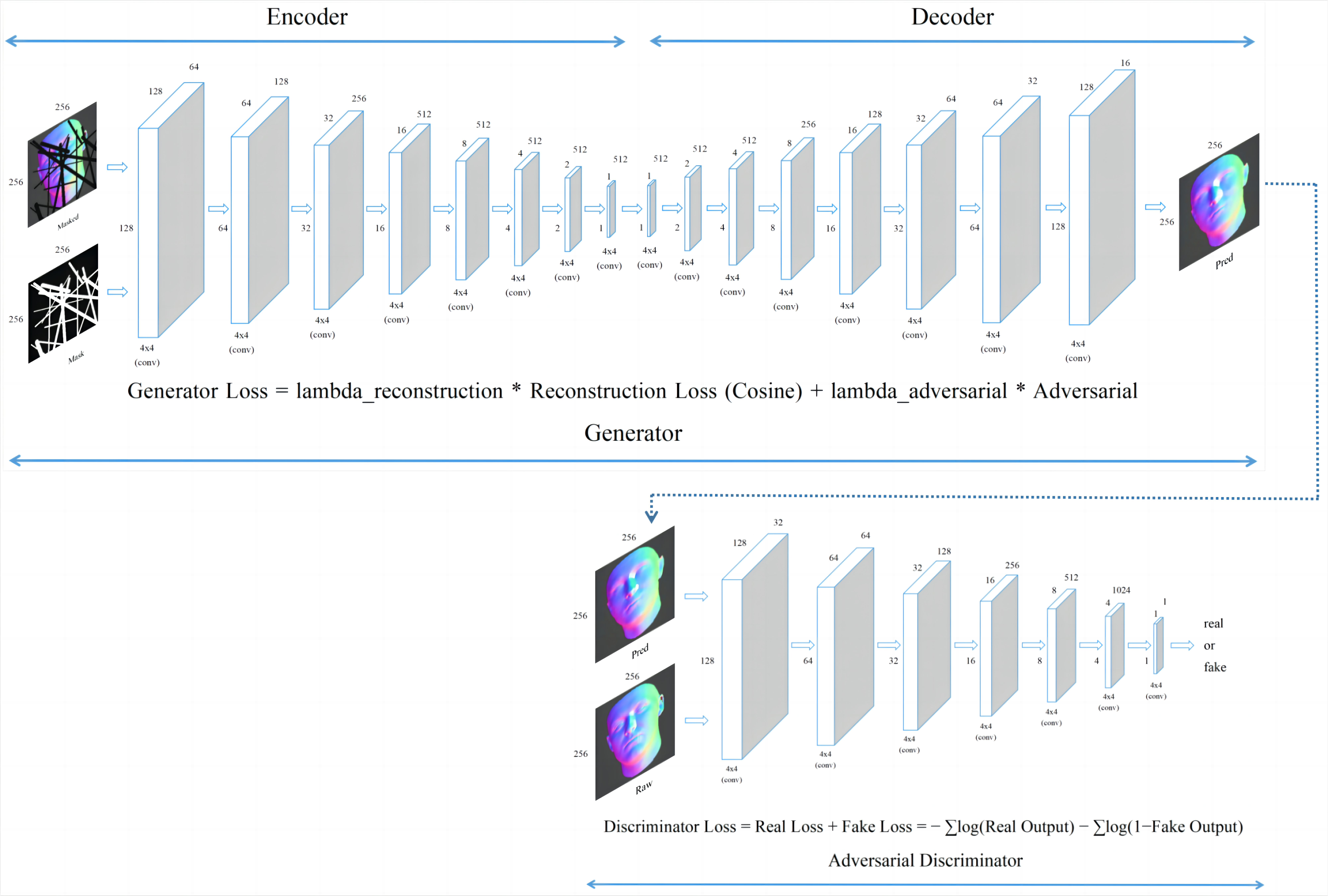}
    \caption{Overview of the GAN model for 256x256 input}
    \label{fig:GAN}
\end{figure*}

\subsection{Model Architecture Selection}

In the context of image inpainting tasks, the Generative Adversarial Network (GAN) model, especially the Deep Convolutional Generative Adversarial Network (DCGAN) variant, offers a powerful and flexible architecture \cite{Radford2015}. Alternative architectures like CycleGAN were considered, but they were found to pose unique challenges in the context of inpainting tasks, such as the difficulty in maintaining spatial consistency and color coherence due to the absence of paired training data \cite{zhu2017unpaired}. Additionally, CycleGAN often requires more computational resources and training time, which may not be feasible in many real-world applications \cite{zhu2017unpaired}. Therefore, due to these complexities and in view of its superior performance in handling image inpainting, the DCGAN architecture was selected.

\subsection{Model Architecture Overview}

The GAN architecture, as shown in Figure \ref{fig:GAN}, consists of a generator, that is intended to take as input corrupted images and produce uncorrupted images as output, and a discriminator that is trained to differentiate between real and fake images \cite{Goodfellow2014}. For inpainting tasks, the generator takes masked images and mask channels as inputs and produces inpainted images. These generated images are then evaluated by the discriminator. The adversarial dynamics between the generator and discriminator facilitates learning of complex, high-dimensional distributions for generating realistic images.

\subsection{Activation Functions and Normalization Techniques}

For the generator model optimized for normal map inpainting, the Leaky ReLU (Rectified Linear Unit) activation function is utilized within its convolutional layers. Leaky ReLU, which allows a small gradient even for negative input values, aids in effective backpropagation, mitigating the "dying ReLU" problem and improving model learning efficiency \cite{Maas2013}.

After the first convolutional layer, batch normalization is applied to stabilize the training process, accelerate learning, and enhance the model generalization ability. By normalizing the activations of the layer, batch normalization minimizes internal covariate shift, an issue where the distribution of layer inputs changes during training, potentially slowing down learning and leading to unstable training dynamics \cite{Ioffe2015}.

To ensure that the generated output represents a normal map, we use a custom layer, the UnitNormalize layer. This layer normalizes the output vectors of the generator to have unit length with respect to the L2 norm. Notably, the normalization operates across the channels of the tensor, preserving the relative contribution of each channel within each pixel.  While the TensorFlow library does include its own keras UnitNormalization layer, we prefer to use a custom layer, to give greater control over aspects such as the treatment of very short / zero length vectors, by including small constant for numerical stability and to prevent division by zero.

\subsection{Optimization Techniques}

The model utilizes the Adam (Adaptive Moment Estimation) optimization algorithm, offering benefits like bias correction and adaptive learning rates \cite{Kingma2014}. Implemented via the Keras API of TensorFlow with a learning rate of \(1 \times 10^{-4}\), it mitigates issues like oscillations and overshooting during the optimization process.

\section{Loss Function}

The loss function for the generator consists of two main components: the reconstruction loss and the adversarial loss \cite{goodfellow2016deep}. The reconstruction loss measures the difference between the target images and the generated images, and the adversarial loss quantifies the ability of the generator to deceive the discriminator.

The reconstruction loss is calculated using the Cosine Similarity between the target images and the generated images. Instead of minimizing the mean squared error, this loss function aims to maximize the sum of scalar products between the generated and ground truth normal vectors across pixels.  In the context of image inpainting, the cosine similarity evaluates the cosine of the angle between the pixel values of the target and generated images, and it can be formulated as:

\begin{equation}
\text{Reconstruction Loss} = 1 - \frac{\sum_{i=1}^{n} y_i \cdot \hat{y}_i}{\sqrt{\sum_{i=1}^{n} y_i^2} \cdot \sqrt{\sum_{i=1}^{n} \hat{y}_i^2}}
\end{equation}

where $y_i$ is the pixel value of the target image, $\hat{y}_i$ is the pixel value of the generated image, and $n$ is the total number of pixels in the image.  Minimizing the above expression will maximize the scalar products.

On the other hand, the adversarial loss is determined using cross-entropy, a widely used loss function for classification tasks \cite{goodfellow2016deep}. In the adversarial context, this cross-entropy measures the ability of the generator to produce images that the discriminator classifies as real, and it can be formulated as:

\begin{equation}
\text{Adversarial Loss} = -\sum_{i=1}^{n} y_i \log(\hat{y}_i) + (1 - y_i) \log(1 - \hat{y}_i)
\end{equation}

where $y_i$ is the true label of the image, and $\hat{y}_i$ is the predicted label by the discriminator.

The total generator loss combines the reconstruction loss and the adversarial loss, with $\lambda_\text{reconstruction}$ and $\lambda_\text{adversarial}$ serving as control weights for each component:

\begin{align}
\begin{split}
\text{Generator Loss} &= \lambda_\text{reconstruction} \cdot \text{reconstruction Loss} \\
&+ \lambda_\text{adversarial} \cdot \text{Adversarial Loss}
\end{split}
\label{eq:generator_loss}
\end{align}

The choice of the lambda parameters $\lambda_\text{reconstruction}$ and $\lambda_\text{adversarial}$, which determine the balance between the importance of each loss component, was based on a brief experimentation with some combinations of integer and non-integer values that were multiples of 10. We found that combinations of integer weights yielded smoother performance and led to integer values when combined. The final settings emphasize the importance of reconstruction loss with $\lambda_\text{reconstruction}$ set to 999.0 and $\lambda_\text{adversarial}$ to 1.0. This specific configuration assists in monitoring changes in loss size during training, facilitating easier analysis without affecting the desired balance between the reconstruction and adversarial loss components. While these values proved effective in our study, a more detailed evaluation might optimize them further.

The generator seeks to minimize this total loss, leading to the generation of images that are visually similar to the target images and able to deceive the discriminator. By optimizing this composite loss function, the generator can produce high-quality inpainted images that are convincing to both the human eye and the discriminator, making it an effective solution for various inpainting tasks and input sizes.

\subsection{Training and Evaluation}

The training process of the Generative Adversarial Network (GAN) model involves an alternating sequence between the generator and the discriminator. During each epoch, multiple batches of data are deployed. The generator receives a batch of masked images and corresponding mask channels with the goal to generate realistic image content in the masked regions. Subsequently, the discriminator is presented with both the real training images and the images generated by the generator for differentiation.

This alternating training procedure ensures continuous improvement in the generator ability to create plausible images and the discriminator proficiency in differentiating real images from those generated. This process undergoes iteration for a predetermined number of epochs, thus enabling the model to effectively hone its image generation and discrimination capabilities.

Upon the conclusion of each epoch, evaluation is conducted using key metrics such as the losses of the generator and discriminator, Structural Similarity Index Measure (SSIM), Peak Signal-to-Noise Ratio (PSNR), and the accuracy of the discriminator. Additionally, visual assessments are executed at regular intervals, providing qualitative insights into the learning progress and the overall performance of the model.

In summary, the model design, comprising the chosen architecture, activation functions, normalization techniques, and optimization algorithm, contributes collectively to the model performance. The iterative and alternating training strategy, coupled with comprehensive evaluation methods, ensures continual refinement of the model performance, thus making it suitable for image inpainting tasks.

\section{Evaluation and Results}

Comparative evaluations are pivotal for understanding the performance of generative models on different masks, elucidating strengths, weaknesses, and potential avenues for improvement. This study contrasts a generative model performance across three masks: the Irregular Lines Mask, Central Blob Mask, and Scattered Smaller Blobs Mask. Simultaneously, the study investigates the influence of the presence or absence of a mask on the model input, particularly focusing on the Irregular Lines Mask.

Metrics such as the Structural Similarity Index (SSIM) \cite{Wang2004}, Peak Signal-to-Noise Ratio (PSNR) \cite{Huynh2008}, and discriminator accuracy serve as primary measures for performance assessment, with all other parameters held constant. This comprehensive evaluation is a significant contribution to the ongoing discourse on generative models, providing valuable insights for researchers aiming to optimize these models for specific masks or to understand the influence of mask inclusion on the inpainting process.

\subsection{Comparative Analysis of Mask Types}

\begin{table}[h!]
\centering
\begin{tabular}{@{}lccc@{}}
\toprule
Mask                  & SSIM (\%)   & PSNR     & Disc Acc (\%)  \\ \midrule
Irregular Lines       & 12.34       & 11.47    & 50.00           \\
Single Big Blob       & 9.47        & 11.44    & 62.86         \\
Scattered Smaller Blobs & 10.98     & 11.43    & 50.00           \\ \bottomrule
\end{tabular}
\caption{Model Performance on Different Masks}
\label{tab:table_lines}
\end{table}

\begin{figure*}[htbp]
  \centering
  \includegraphics[width=0.77\linewidth]{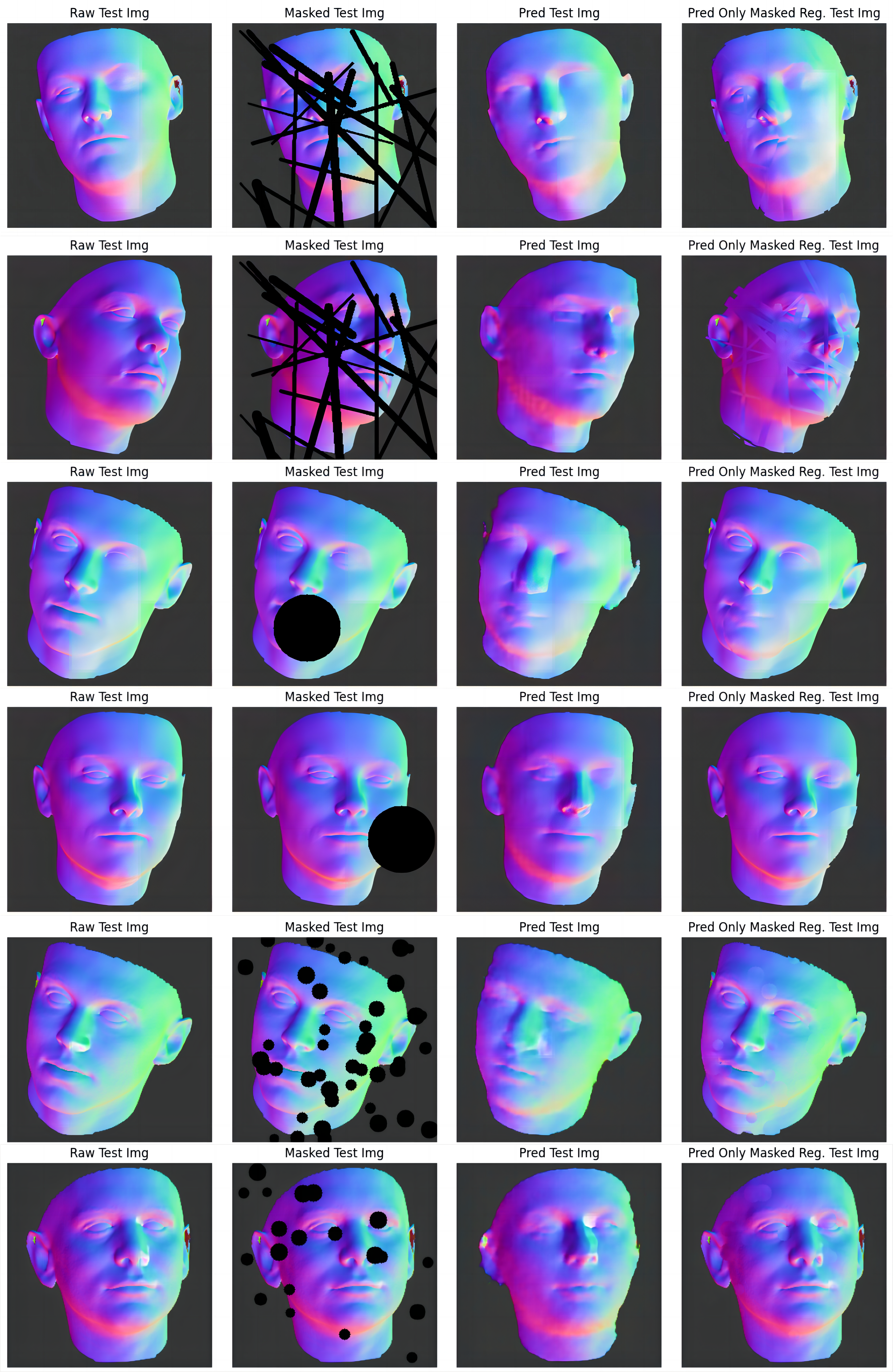}
  \caption{From top to bottom Performance on Irregular Lines,  Single Big Blob and Scattered Smaller Blobs Masks; From Left to Right compares of the Raw Image, Masked Image, Predicted Image, and Predicted Image in the Masked Region Only}
  \label{fig:performance}
\end{figure*}

Although the metrics demonstrate varied results across the three masks, the visual evaluation of the generated images (as depicted in Figure \ref{fig:performance}) indicated superior performance with the Single Big Blob Mask. Despite a lower SSIM score (9.47\%) and a similar PSNR value (11.44) compared to other masks, this mask yielded the most visually satisfying inpainting outcomes. The discriminator accuracy for the Single Big Blob Mask was higher (62.86\%), suggesting that the generator still has some difficulty in perfectly deceiving the discriminator, potentially indicating room for further model optimization.

The Irregular Lines Mask resulted in the highest SSIM (12.34\%), yet this did not translate into superior visual results. The PSNR value was slightly higher than for other masks, but within the same range. Discriminator accuracy for this mask was at 50\%, indicating a balance between the generator and discriminator.

Performance with the Scattered Smaller Blobs Mask produced a balanced SSIM score (10.98\%) and a similar PSNR value (11.43) to the other masks. However, the discriminator accuracy mirrored that of the Irregular Lines Mask, again indicating an equilibrium between the generator and the discriminator.

These findings provide critical insights into model performance across different masks and suggest potential directions for future research. Further refinement to the model may include hyperparameter tuning or architectural adjustments to accommodate specific mask characteristics. While metrics such as SSIM and PSNR provide valuable quantitative evaluations, the assessment of inpainting models should be holistic, balancing these with visual evaluations, as observed in the case of the Single Big Blob Mask.

\subsection{Impact of Mask Presence on Model Input}

\begin{table}[htbp]
\centering
\begin{tabular}{@{}lccc@{}}
\toprule
Mask                  & SSIM (\%)   & PSNR     & Disc Acc (\%)  \\ \midrule
With       & 12.34       & 11.47    & 50.00           \\
Without    & 11.50       & 11.48    & 50.00            \\ \bottomrule
\end{tabular}
\caption{Model Performance With and Without Mask Input Training}
\label{tab:table_mask}
\end{table}

\begin{figure}[htbp]
  \centering
  \includegraphics[width=1.0\linewidth]{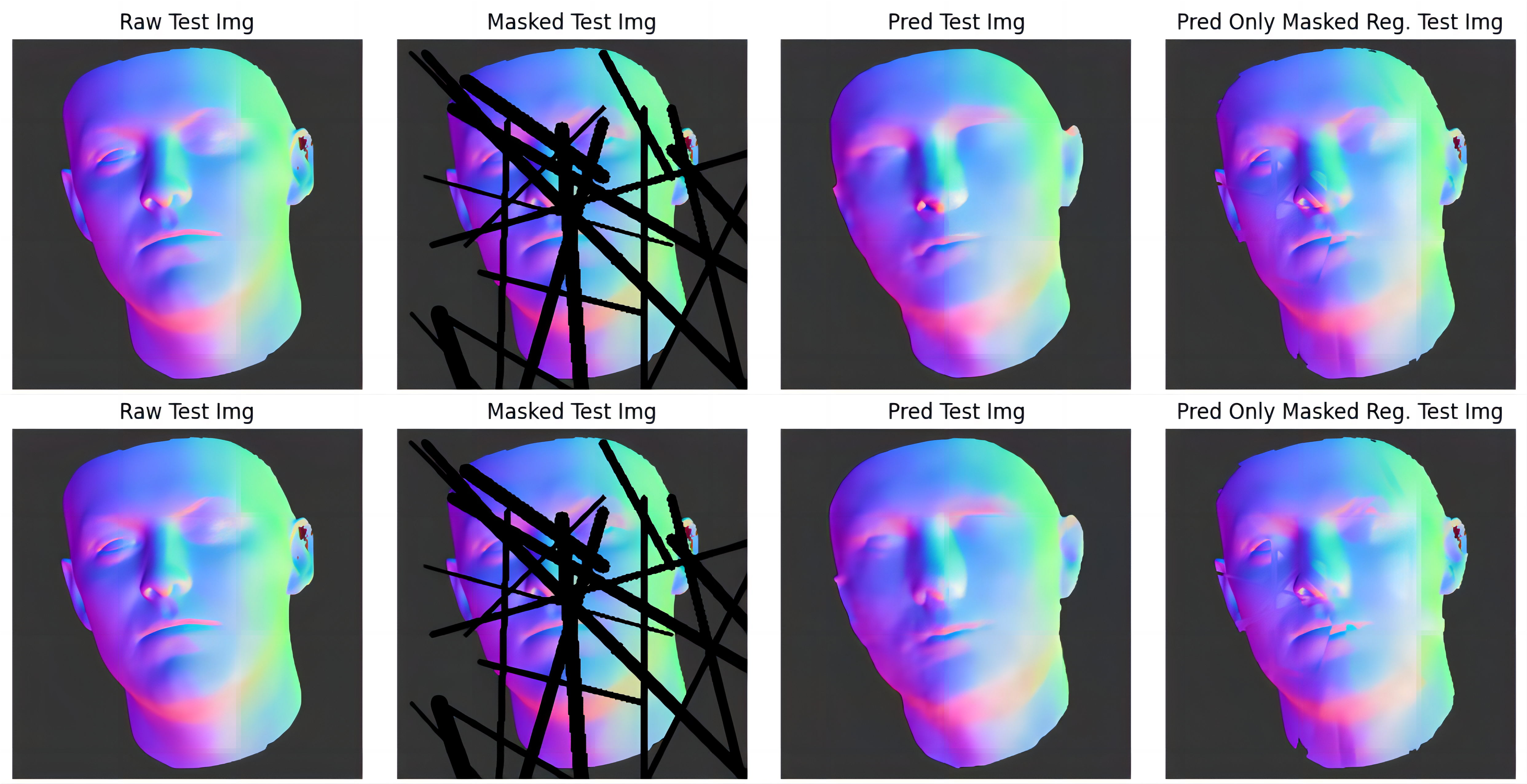}
  \caption{From top to bottom Performance on With and Without on Irregular Lines Mask; From Left to Right compares of the Raw Image, Masked Image, Predicted Image, and Predicted Image in the Masked Region Only}
  \label{fig:withWithMask}
\end{figure}

A critical aspect of this study was the examination of the generative model performance when trained with and without the presence of a mask in the model input, with a particular focus on the Irregular Lines Mask. As observed from Table \ref{tab:table_mask}, the Structural Similarity Index (SSIM), Peak Signal-to-Noise Ratio (PSNR), and the discriminator accuracy metrics reveal that there is no significant discrepancy in performance with the mask presence. Both approaches yield comparable results, with the SSIM marginally higher (12.34\%) when the mask is included, and the PSNR values being almost identical. 

However, a closer inspection of the inpainted images in Figure \ref{fig:withWithMask} provides a different perspective. Although at first glance, the differences may seem negligible, the images where the mask channel was incorporated during the generator training demonstrate a slightly more accurate rendering of details in certain areas. 

This subtle improvement, though not largely quantified by the metrics, highlights the potential benefits of using mask input in training. Even minor enhancements in detail accuracy can contribute to the overall realism and visual quality of the generated images, underpinning the value of a more comprehensive and nuanced evaluation approach that incorporates both quantitative metrics and qualitative visual assessments.

These results suggest that the introduction of the mask channel in the generator training phase can potentially refine the performance of the generative model. However, it also highlights the need for further research to optimize the use of mask input, particularly regarding how different mask types and structures may interact with various training configurations.

\section{Conclusions}

Through rigorous assessment across three different masks: Irregular Lines Mask, Central Blob Mask, and Scattered Smaller Blobs Mask, the performance of the inpainting model was carefully evaluated. This wide-ranging evaluation helped gain insights into how the model performed across different masking scenarios and identified the nuanced strengths and improvement areas of the model.

In the visual assessment of inpainted images, it was observed that the model generated satisfactory and visually pleasing results. The integration of the mask channel and the expansion of the layer structure, while relatively minor modifications, contributed to these improved results.

From the conducted experiments and derived conclusions, it is evident that the GAN-based inpainting model has been successful in addressing the set objectives, albeit with modest improvements. The implementation of a mask channel in the input and the expansion of the layer structure resulted in a perceptible enhancement in the model performance, leading to better quality inpainted images. The comprehensive testing across various masks underscores the adaptability of the model, suggesting potential for wider applications in the image inpainting domain.

\section{Future Work}

This research has explored promising directions in image inpainting, but there is still room for further improvement and expansion.

Alternative network architectures such as the U-Net style architecture \cite{ronneberger2015unet} could be explored to possibly yield more high frequency features for the reconstruction phase, thereby potentially enhancing the quality of the inpainted images.

A further research direction could involve a deeper investigation into the integrability of the predicted normal maps. Evaluating whether there could be a surface whose normals are consistent with the given normal map could serve as an additional validation of the generated maps. The method outlined in "Normal Integration via Inverse Plane Fitting With Minimum Point-to-Plane Distance" CVPR 2021 \cite{Cao_2021_CVPR} offers one potential approach to this investigation.

Moreover, evaluating the human-like appearance of the predicted normal maps could provide another meaningful evaluation metric. This could include assessing elements such as the reconstruction of nostrils and the continuity of lips. The use of the normal map for relighting scenarios and the subsequent evaluation of its performance in these settings would provide a practical measure of the quality of the map.

There is also potential for further optimization in the balance of the reconstruction and adversarial loss components. While integer-based weights were selected in this study due to their observed smoother performance, a more extensive evaluation of different weightings could lead to a more optimal balance and improve the overall performance of the model.

Finally, the inclusion of more diverse, and potentially more perceptually-oriented metrics could improve the evaluation of the quality of the results, expanding beyond the current evaluation metrics utilized in this study.

By pursuing these lines of inquiry, we can continue to refine and advance the capabilities of image inpainting technology.


\bibliographystyle{eg-alpha-doi} 
\bibliography{egbibsample}       



\end{document}